%% file: main.tex
\documentclass[conference]{IEEEtran}
\IEEEoverridecommandlockouts
\usepackage{cite}
\usepackage{amsmath,amssymb,amsfonts}
\usepackage{algorithmic}
\usepackage{graphicx}
\usepackage{textcomp}
\usepackage{xcolor}
\usepackage{caption}
\usepackage{subcaption}
\def\BibTeX{{\rm B\kern-.05em{\sc i\kern-.025em b}\kern-.08em
    T\kern-.1667em\lower.7ex\hbox{E}\kern-.125emX}}
\begin{document}

\title{\Large \bf Autonomous Shuttle Operation for Vulnerable Populations: Lessons and Experiences}

\author{Ren Zhong$^{1}$, Zhaofeng Tian$^{2}$, Jinghui Liao$^{1}$ and Weisong Shi$^{2}$ \\
{The CAR Lab, University of Delaware} \\

{\tt Technical Report: CAR-2024-001}
\thanks{$^{1}$Ren Zhong and Jinghui Liao are with the Department of Computer Science, Wayne State University.
        % {\tt\small liangkai@wayne.edu, weisong@wayne.edu}
        }%
\thanks{$^{2}$Weisong Shi and Zhaofeng Tian are with the Department of Computer and Information Sciences, University of Delaware.
        % {\tt\small b.d.researcher@ieee.org}
        }%
}

\maketitle

\begin{abstract}
\label{section:abstract}
\input{0_Abstract}
\end{abstract}


\section{Introduction}
\label{section:introduction}
\input{1_Introduction}

\section{Operation Design}
\label{section:experimentdesign}
\input{2_ExperiementDesign}

\section{Operation Observations}
\label{section:experimentResult}
\input{3_ExperimentResults}

\section{Public Shuttle Planning Design Requirement}
\label{section:planningrequirement}
\input{4_planningRequirements}

\section{Discussion}
\label{section:discuss}
\input{5_Discussion}

\section{Summary}
\label{section:summary}
\input{6_Summary}

\section{Acknowledgement}
\label{section:Acknowledgement}
\input{7_Acknowledgement}

\bibliographystyle{IEEEtran}
\bibliography{main}

\end{document}

%% file: 0_Abstract.tex
The increasing shortage of drivers poses a significant threat to vulnerable populations, particularly seniors and disabled individuals who heavily depend on public transportation for accessing healthcare services and social events. Autonomous Vehicles (AVs) emerge as a promising alternative, offering potential improvements in accessibility and independence for these groups. However, current designs and studies often overlook the unique needs and experiences of these populations, leading to potential accessibility barriers. This paper presents a detailed case study of an autonomous shuttle pilot specifically tailored for seniors and disabled individuals, conducted during the early stages of the COVID-19 pandemic. The service, which lasted 13 weeks, catered to approximately 1500 passengers in an urban setting, aiming to facilitate access to essential services. Drawing from the safety operator's experiences and direct observations, we identify critical user experience and safety challenges faced by vulnerable passengers. Based on our findings, we propose targeted initiatives to enhance the safety, accessibility, and user education of AV technology for seniors and disabled individuals. These include increasing educational opportunities to familiarize these groups with AV technology, designing AVs with a focus on diversity and inclusion, and improving training programs for AV operators to address the unique needs of vulnerable populations. Through these initiatives, we aim to bridge the gap in AV accessibility and ensure that these technologies benefit all members of society.

%% file: 1_Introduction.tex
Public transportation has become less integral to daily life for many since the advent of affordable personal automobiles. This shift has led to an underestimation of public transportation's value, particularly ignoring the essential role it plays for over three million seniors and disabled individuals \cite{elderlytransportation}  in the United States. These groups rely on public transit for essential activities such as employment, healthcare, and community participation, which are crucial for their quality of life.  However, these vulnerable populations face significant barriers in accessing public transportation, primarily due to budget constraints that limit service availability and inconvenient fixed routes from designated stations. Additionally, a report by TransitCenter \cite{busdrivershortage} highlights a crisis: a significant portion of U.S. bus operators are aged between 45 to 64, suggesting potential workforce shortages of up to 200,000 drivers in the near future.

To address the mobility needs of seniors and the disabled, governments since 1973 have supported paratransit, a service offering individualized rides without fixed routes or schedules, at a cost lower than taxis or ride-hailing services, and equipped with accessibility features. However, the effectiveness of paratransit is limited by the need for advance booking. Additionally, the shortage of drivers problem will also negatively impact paratransit. 

In recent years, with the evolution of sensing technology and computer vision algorithms, Autonomous Vehicles (AVs) have been made available to the public. Robotaxis from NuTonomy was the first company starting to offer taxi rides in 2016. Since then, Waymo, GM, Ford, etc. have started their own robotaxis services. Vulnerable people can set accurate pick-up and drop-off locations for a travel request through a mobile app. Then, the driving system of AVs can dynamically plan driving routes that offer either minimum waiting time or lower travel prices. However, as a desired solution for improving vulnerable people's transportation, these robotaxis are mostly developed based on normal-size cars without wheelchair lifts or ramps. Companies such as NAVYA and EasyMile have noticed this problem and are developing self-driving shuttles to provide sufficient accessibility. These shuttles don't have a powerful driving performance like robotaxis to travel freely on the road, thus focusing on accomplishing the First Mile/Last Mile tasks in fixed areas, such as communities, hospitals, campuses, etc.. Given these limitations, the question of whether vulnerable populations can benefit from this cutting-edge technology remains a significant gap in our understanding.

Existing research has explored public acceptance, trust, and behavior towards autonomous shuttles through road pilots, with notable experiments conducted in environments like Mcity. These studies provide valuable insights into the operational and engagement aspects of deploying autonomous shuttles. However, the relevance of findings from settings like Mcity to vulnerable populations is limited, as these areas are not typically frequented by seniors or disabled individuals. This paper aims to bridge this gap by detailing a new deployment of autonomous shuttles in communities with a significant presence of vulnerable residents, focusing on providing transitional services among essential locations such as hospitals. In summary, this paper makes the following contributions:

\begin{itemize}
    \item We present comprehensive details of the first public road pilot for Autonomous Vehicles (AVs) specifically aimed at enhancing transportation accessibility for seniors and disabled individuals. This pilot focuses on facilitating their access to healthcare services and social events, aiming to fill the gap in current transportation solutions. 
    \item Throughout the testing phase, we identified numerous challenging scenarios that test the limits of autonomous driving technology. In response to these edge cases, we propose four research initiatives aimed at advancing research in the field of autonomous driving, with a particular focus on improving safety and reliability for vulnerable populations.
\end{itemize}

The remainder of this paper is structured as follows: Section~\ref{section:experimentdesign} outlines the detailed plan and methodology of the autonomous shuttle pilot. Section~\ref{section:experimentResult} discusses the observations and data collected by the safety operator during the testing period. Section~\ref{section:discuss} introduces four research initiatives aimed at enhancing autonomous vehicle technology, derived from insights gained during the pilot. Finally, Section~\ref{section:summary} provides a conclusion to the paper, summarizing the key findings and contributions.

%% file: 2_ExperiementDesign.tex
% Generally, the test for autonomous driving vehicles requires consideration of the Operational Design Domain (ODD) \cite{operationaldesigndomain} including four aspects, speed, environment, weather, and traffic patterns. 
In this section, we present a detailed test plan in four aspects, vehicle specifications, route and time schedule, deployment communities, and safety operators. 

\begin{figure}[t]
	\centering
	\includegraphics[width=0.8\columnwidth]{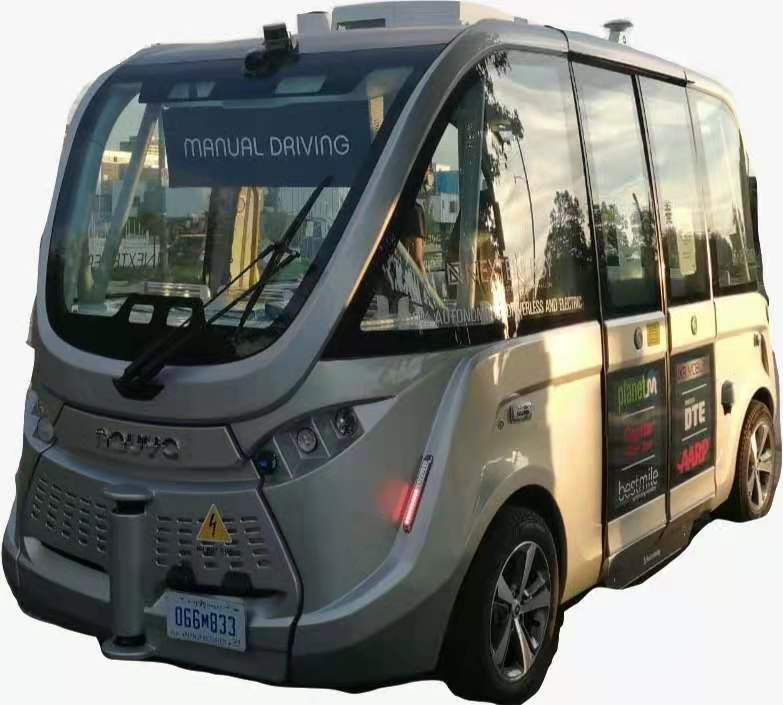}
	\caption{Test Shuttle Appearance}
	\label{fig:shuttle}
\end{figure}

\subsection{Vehicle Specifications}

Fig. \ref{fig:shuttle} shows the exterior of the autonomous shuttle, designed for steering challenge scenarios by allowing four-wheel drive and rotation capabilities. The shuttle is powered by an 80-volt battery, supporting nine hours of continuous operation at 25 km/h, although the driving speed is capped at 18 km/h for safety reasons. It is equipped with eleven seats and two standing areas, accommodating up to thirteen passengers, including a safety operator. The shuttle features a uniform sensor configuration on both sides, including a radar, a camera, and a LiDAR, for environmental awareness. Additionally, a high-accuracy RTK-GNSS receiver is mounted on the top for precise localization."

Within the shuttle, a specialized computing device is implemented to manage the autonomous driving system, which is responsible for executing various driving tasks. Among components, the navigation system and the localization system are two core modules of the autonomous driving system. They are used to track the vehicle's movement, plan safe routes, and execute the driving plan. The shuttle's localization system fuses three different technologies to ensure robust tracking of vehicle movement. 

\begin{itemize}
\item \textbf{RTK-GNSS Positioning:} Due to satellite connection uncertainty, the Global Navigation Satellite System (GNSS) achieves only meter-level accuracy in the localization layer. However, Real-Time Kinematic GNSS (RTK-GNSS) enhances localization performance to centimeter-level accuracy. This improvement is accomplished by connecting to fixed RTK stations, which continuously transmit corrections for satellite signals. These corrections are based on the stations' known positions and are transmitted via mobile networks. Currently, there are two types of Real-Time Kinematic (RTK) stations: public stations and privately built self-operated stations. Due to the absence of public stations in the vicinity of the test area, a local RTK base station was deployed to ensure a consistent and stable connection.

\item \textbf {Odometer Tracking:} The odometer, which tracks vehicle movement by measuring the rotation speed of the tires and the steering angle to calculate the vehicle's speed and orientation, is the most commonly used solution. However, the inherent measurement errors of the tire rotation sensor, coupled with uneven road surfaces, can lead to drifted calculation results. Consequently, the odometer is primarily utilized for short-term estimation of vehicle movement.

\item \textbf{Lidar-based Tracking:} Lidar-based tracking is widely utilized in autonomous robots for localization and tracking, as it can provide accurate spatial coordinates. This accuracy is achieved by finding the best match between real-time data and the pre-built map. These coordinates are essential for the navigation system to plan the subsequent route to the target. In autonomous vehicles that have sufficient space and power supply, high-accuracy 3D Lidar systems can be equipped to provide more reliable localization, as long as the environment remains largely unchanged. Unfortunately, public roads can receive changes due to various factors, such as new construction zones and parked vehicles. These changes result in discrepancies between the online sensor data and the pre-existing map, causing the matching process to fail.

\end{itemize}

By jointly utilizing these localization methods, the accuracy achieved is sufficient for the navigation system to operate successfully. The localization information is referenced on a High Definition (HD) map\cite{yang2018hdnet} which is built during the preparation phase of the test. This construction involves manually driving a shuttle around communities to collect sensor data; notably, LiDAR-scanned point clouds of roads are recorded. These point clouds are essential for SLAM algorithms, such as those described in \cite{ndtforautoware}, which merge the separate point clouds into a comprehensive localization layer of the HD map. Above this layer, map editors develop the semantic layer that illustrates the driving paths within the testing region. Additionally, they align a geographic map obtained from OpenStreetMap \cite{OpenStreetMap}, enabling the integration  of GNSS coordinates into the localization layer.

% As a self-driving car, the shuttle must be able to detect objects on the road to avoid collisions. Since then, the final method of locating the shuttle is to use Lidar SLAM technology. This involves comparing the stored map with the map seen by the shuttle in real-time. To this end, the shuttle is equipped with a 3D Lidar on the top of the shuttle to get a 3D perception of the environment and detect surrounding obstacles in real-time. It also has cameras that can capture the visual environment in the event of an accident.

\begin{figure}[htbp]
    \centerline{\includegraphics[width=0.76\columnwidth]{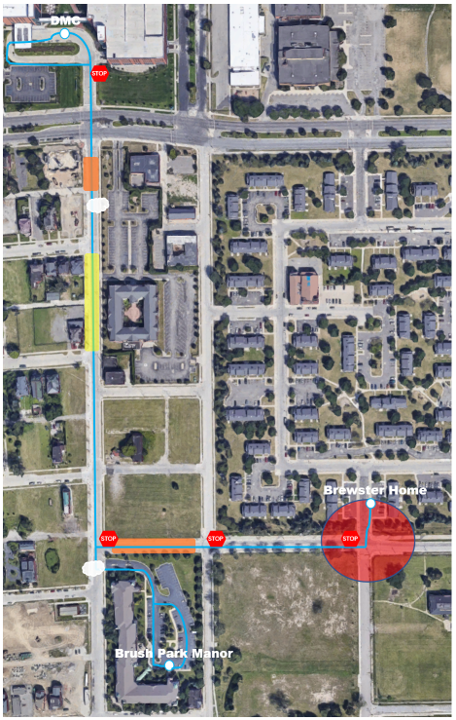}}
    \caption{Operation Route and Special Event Area}
    \label{fig:routemap}
\end{figure}

\subsection{Test Specifications}

The primary objective of this test was to evaluate the acceptance and accessibility of autonomous vehicles among seniors and disabled individuals. To this end, the route was established between two residential communities and a medical center. This was designed to meet the residents' needs for accessing medical services. Additionally, the route connecting the two communities aimed to facilitate attendance at social events, addressing another critical aspect of the customers' requirements.

\subsubsection {Operation Route}

The blue line in Fig. \ref{fig:routemap} illustrates the operation route, which is predetermined and operates on a periodic basis to serve the designated stops. Brush Park Manor serves as the location where the shuttle is parked and charged overnight. The route encompasses an urban operational domain, segmented into distinct scenarios: parking lots, residential communities, T-intersections, four-way intersections, and hospital entrances. The route also includes several stop signs that the shuttle is programmed to recognize and respond to. Additionally, Fig. \ref{fig:routemap} illustrates special event areas, previously unexpected, with different color blocks: white indicates areas where steam was frequently present in the winter; orange denotes construction zones; yellow signifies zones where cars were often parked on roadside and block the road; and red highlights areas where RTK-GNSS receiver failed to connect RTK base station frequently. These events can cause the shuttle to shut down unexpectedly. We will introduce these events in Section \ref{section:experimentResult}.

\subsubsection {Time Schedule and Weather Conditions}

The testing schedule was set for five days per week, with operations starting at 8:30 AM and ending after 5:30 PM. Each service, following the route described earlier, lasted approximately 30 minutes. In response to COVID-19 restrictions, the maximum number of passengers was limited to two. Consequently, as the presence of a safety operator on the shuttle is mandatory, only one additional passenger was allowed per trip. Accounting for a one-and-a-half-hour break for the operator, the shuttle could accommodate up to 45 passengers each day, assuming each passenger disembarked at the next stop. The shuttle was capable of operating in various weather conditions autonomously without operator intervention, except during rain, fog, and snow. However, for safety reasons, testing was conducted only on sunny and cloudy days. The test period spanned 13 weeks, starting from August 3rd and concluding on October 28th, terminated early due to the onset of winter.

\begin{figure}[htbp]
\centerline{\includegraphics[width=0.6\columnwidth]{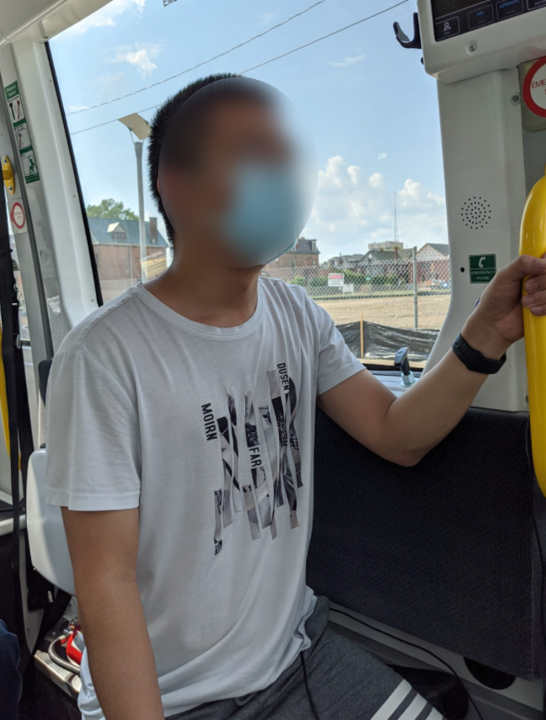}}
\caption{Operator Focus On Road}
\label{fig:shuttleoperator}
\end{figure}

\subsection{Safety Operator}

To ensure safety during each test, an operator is employed to assume control of the shuttle in instances of autonomous driving system failure. Fig. \ref{fig:shuttleoperator} depicts an operator attentively monitoring the road conditions. Additionally, the shuttle is equipped with a controller, allowing the operator to manually maneuver the vehicle as necessary.

%% file: 3_ExperimentResults.tex
During the testing phase, while the shuttle was successfully operated autonomously most of the time, operators still encountered various safety-critical issues due to unexpected system malfunctions. As the test progressed, operators accumulated experience and identified patterns in these safety issues, which enabled them to proactively take manual control of the vehicle to prevent potential incidents. In this section, as the shuttle manufacturer refuses to disclose the operational data, we will focus on presenting observations made by operators during the operation. This includes issues related to safety and passengers' experiences during the ride.

\subsection{Emergence Brake Issues}

Safety operators had to maintain full concentration on maintaining vigilance during autonomous shuttle operations as unexpected issues can cause discomfort and even worse safety issues. A primary source of discomfort arises from the autonomous driving system's tendency to engage in sudden emergency braking without prior warning, often due to the loss of RTK-GNSS signals among other factors.  Such disruptions in GNSS signals can lead to significant positional inaccuracies, forcing the autonomous system to execute immediate braking to avoid deviations from the intended route. Notably, GNSS signal loss occurs unpredictably, with a marked increase in frequency in bad weather. This abrupt braking can result in passengers being jolted forward, particularly if they are not secured with seat belts. For example, an incident was recorded where a passenger, despite gripping a support ring while standing to observe the scenery, was thrown off balance by an emergency stop resulting in minor injuries. Similarly, the following cases can also cause uncomfortable experiences:

% to promptly identify and address safety concerns The most significant cause of discomfort was the autonomous driving system's sudden emergency braking without warning, often resulting from GNSS signal loss among other factors. This loss of GNSS signal could lead to substantial positional inaccuracies, prompting the autonomous system to initiate immediate braking to prevent the shuttle from deviating onto an incorrect route. GNSS signal loss tended to occur randomly, with a higher incidence on cloudy days. Such unexpected braking could cause passengers to lurch forward if they were not properly secured by seat belts. In one instance, despite holding onto a hanging ring while standing to view the scenery, a passenger lost balance during an emergency stop triggered by GNSS signal loss and sustained minor injuries. This underscores the importance of proper safety precautions, such as seat belt usage, during autonomous shuttle operation. In addition, there are situations that can cause similar uncomfortable experiences, including:

\begin{itemize}
\item \textbf{Path Planning Failure.} After manually intervening to navigate past safety concerns, the operator typically halts the shuttle close to the intended route before reactivating autonomous mode. Generally, the autonomous driving system is capable of resuming and completing the designated route. However, malfunctions in the navigation system's path planning can occur, leading to the generation of incorrect routes. If the shuttle deviates significantly from the correct path, it will execute an emergency brake. In a severe instance, the shuttle suddenly swerved nearly 45 degrees, nearly throwing the operator off his seat. 

% such deviations can prompt the shuttle to make an abrupt maneuver, turning approximately 70 degrees, which subjects the operator to a violent shake. This underscores the critical need for reliable navigation and smooth transition mechanisms between manual and autonomous modes to ensure the safety and stability of the shuttle's operations.

\item \textbf{Detection Failure Caused Emergency Brake.} The steam pouring out of certain manholes during cold seasons is a common occurrence in the test area. When the shuttle encounters areas with intermittent steam, its behavior is unpredictable due to the steam's interference. Although the steam is harmless fog, the shuttle's lidar-based obstacle detection system may mistakenly identify it as a tangible obstacle, prompting the system to apply emergency brakes to avoid a collision. In certain instances, this can result in the shuttle repeatedly accelerating and braking, requiring operator intervention to take manual control and ensure safe navigation through the steam region.
\end{itemize}

% The situations described above are the results of a safety assurance mechanism that can avoid severe collision accidents as much as possible when passengers have fastened their seat belts. In normal situations, the shuttle will slow down gradually when obstacles on the route are detected.

\begin{figure}[htbp]
     \centering
     \begin{subfigure}[b]{0.36\textwidth}
         \centering
         \includegraphics[width=\textwidth]{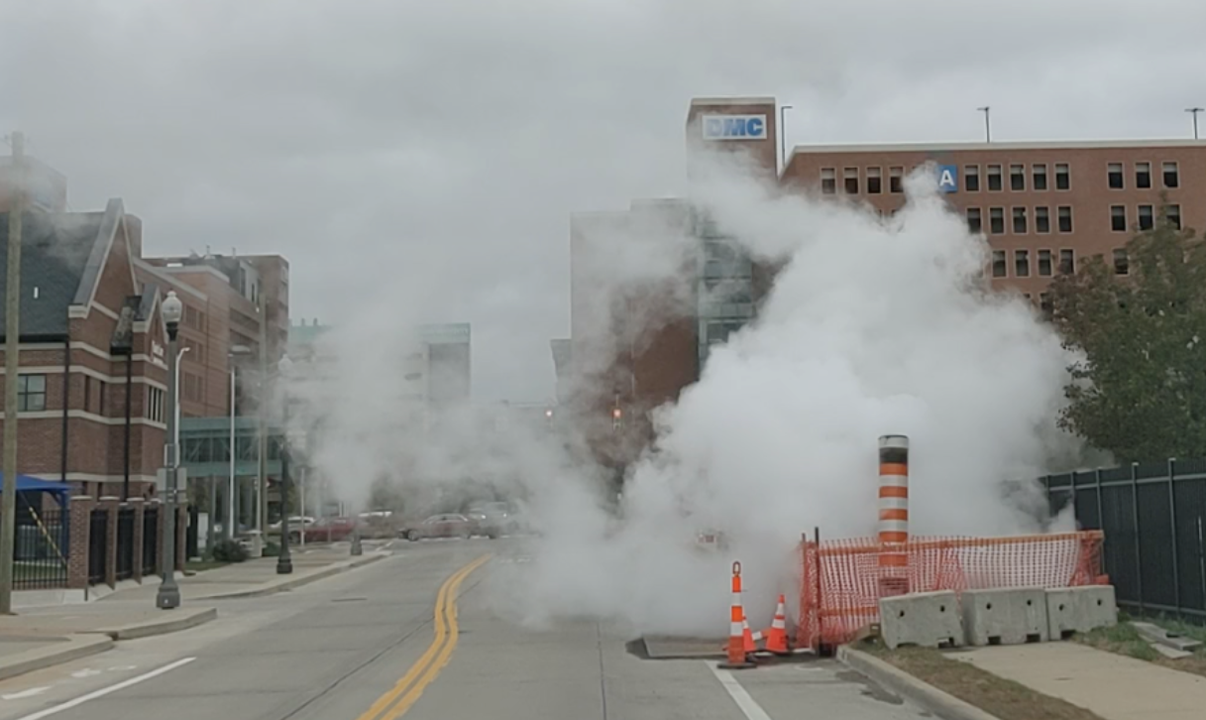}
         \caption{Heavy steam from roadside}
         \label{fig:steam}
     \end{subfigure}
     \hfill
     \begin{subfigure}[b]{0.36\textwidth}
         \centering
         \includegraphics[width=\textwidth]{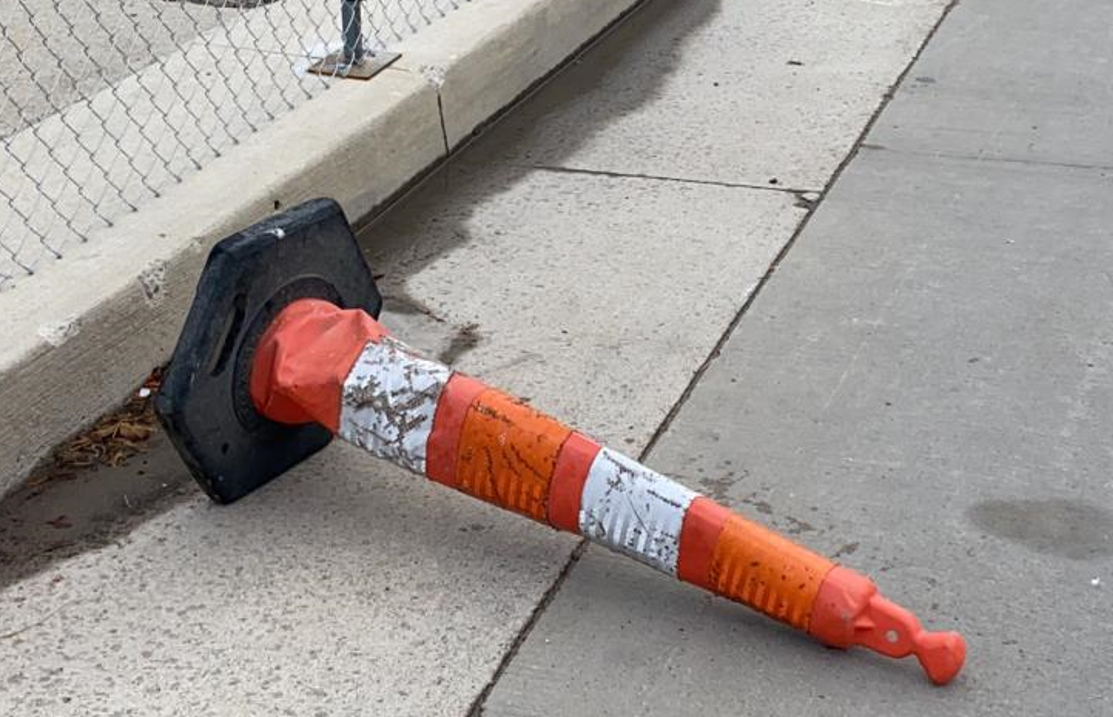}
         \caption{Cone on ground in blind spot}
         \label{fig:cone}
     \end{subfigure}
        \caption{Examples of unpredictable obstacles}
        \label{fig:obstacles}
\end{figure}

\subsection{Uncommon Obstacles Issues}
Besides the obstacles that obviously blocked the route, such as the sign of the work zone and vehicles parked at the roadside, there were some unpredictable objects that did not block the shuttle but influenced the Lidar-based obstacle detection. The steam is a product of the heating system, which relies on the tube in each region to discharge the steam during the winter. The cameras and Lidar sensors can both sense the existence of the steam; this is also treated as an obstacle, as are branches of some trees and discards on the roadside that do not block the road. A rigorous obstacle detection system can improve safety but degrade passengers' riding experiences as the shuttle will keep waiting until obstacles disappear. Another case is a work zone cone that fell to the ground. The system correctly sensed its existence and braked before crushing the cone. However, the cone seemed to be in the blind spot of the detection system, so the shuttle resumed moving forward and then crushed the cone. Fig. \ref{fig:obstacles} shows these two unpredictable obstacles.

\subsection{Autonomous Driving Issues}
The autonomous shuttle is operated along a fixed route that was defined before the operation started. In this manner, autonomous vehicles only need to focus on positioning and detecting obstacles without planning routes in real time. The test shows that the autonomous shuttle can operate safely and smoothly most of the time in urban areas. However, we also observed that some scenarios are challenging for this kind of system to handle. 

\begin{figure}[htbp]
     \centering
     \begin{subfigure}[b]{0.42\textwidth}
         \centering
         \includegraphics[width=\textwidth, height=5cm]{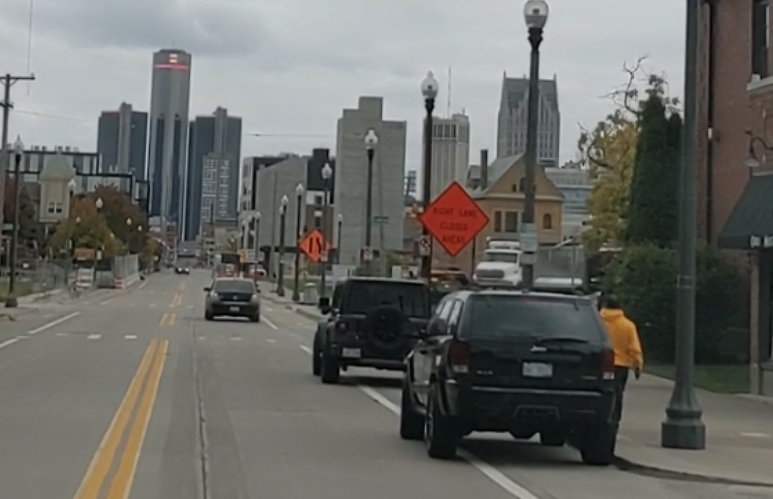}
         \caption{Parked Cars Block Road}
         \label{fig:roadsidepark}
     \end{subfigure}
     \hfill
     \begin{subfigure}[b]{0.42\textwidth}
         \centering
         \includegraphics[width=\textwidth, height=5cm]{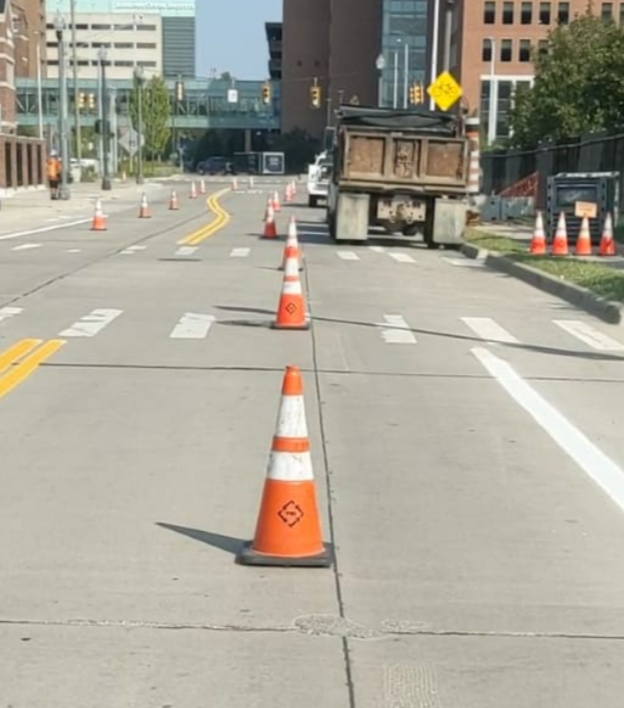}
         \caption{Work zone block road}
         \label{fig:workzone}
     \end{subfigure}
     \hfill
     \begin{subfigure}[b]{0.42\textwidth}
         \centering
         \includegraphics[width=\textwidth, height=5cm]{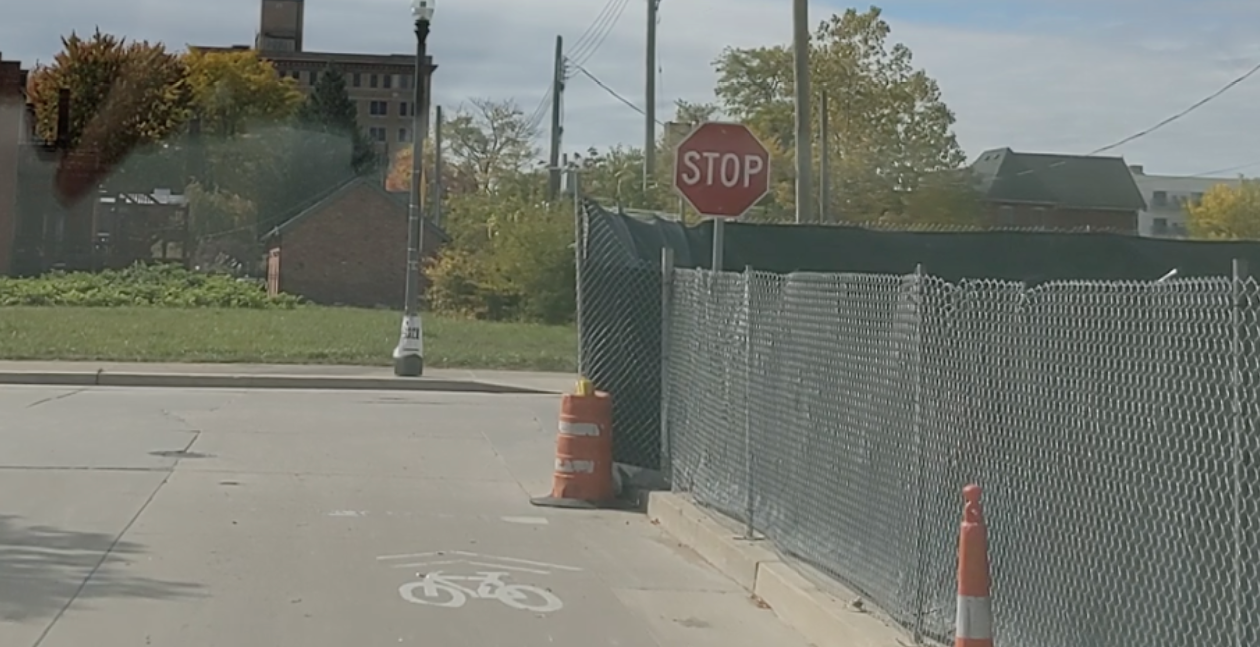}
         \caption{Construction zone block view}
         \label{fig:wallblockview}
     \end{subfigure}
    \caption{Obstacle influence autonomous driving}
    \label{fig:bigobstacles}
\end{figure}

\begin{itemize}
\item \textbf{Crossing Intersections.} Since the shuttle requires accurate localization information to keep following the route, when localization is difficult, safety operators have to manually control vehicles to avoid incidents. A four-way intersection is a challenging scenario where a Lidar-based localization system can't match sufficient features. For some other scenarios, the operator doesn't need to fully control the shuttle. For example, the shuttle will fully stop to ensure safety. However, the shuttle can't resume moving automatically, as it can't comprehend the behavior of vehicles from the side direction so the operator needs to decide the time the shuttle restart to move.

\item \textbf{Obstacle Blocks Route.} The urban environment is dynamic, and road maintenance and construction works may significantly change the environment by adding road signs and obstacles. The shuttle is not capable of bypassing the obstacle. As a result, the shuttle always waits until the obstacle is removed. The design is based on two reasons. The first is that the shuttle operates on a fixed route without considering real-time path planning. Another is that due to the limited perception range of the sensors, the shuttle may run into a car coming from another lane when bypassing the obstacle on the narrow urban road. In Fig. \ref{fig:bigobstacles}, we present several examples of obstacles that block the route. The first two are the vehicles parked on the roadside and the work zone. The last one shows the wall of the construction area. Although the wall did not block the route, the right side of the field of view is completely occluded. Although we did not meet any real accidents during the test, the operator was scared when a vehicle suddenly appeared while he was trying to turn right at the corner.

\item \textbf{Resume Autonomous Driving.} Every time the operator needs to manually drive the shuttle, he also needs to resume the operation mode back to autonomous. However, the path planning capability of the shuttle is not sufficient to drive itself back to route at any status. The operator should ensure that the shuttle is within three meters of the route, that the angle between the head of the shuttle approaching the route is less than three meters, and that the angle between the facing direction of the shuttle and the route is within 10 degrees. This process may fail when the localization system is not working and the operator does not realize it. The failure causes emergency brakes or sharp turns, which are unsafe for both passengers and other road occupations.

\end{itemize}

\subsection{Feedback from Safety Operators and Passengers}

During the 13-week testing period, the service catered to approximately 1,500 passengers across at least 2,800 instances, employing four operators to guarantee safety and maintain vehicle sanitation. Feedback from both operators and passengers has been compiled in this section to assess the overall experience and identify areas for improvement.

Safety operators faced significant stress due to the previously mentioned safety concerns, necessitating constant vigilance and decision-making on when to manually override the autonomous system. Despite their crucial role, the operators were not provided with a comfortable working environment. As depicted in Fig \ref{fig:shuttleoperator}, operators had to focus intently on the road while sitting in a backup seat without a backrest, making long periods of observation and operation physically taxing. The seatbelt was not designed for this unconventional seating posture, adding to their discomfort. Critical vehicle information, such as GNSS connectivity and sensor data, was displayed on a terminal positioned inconveniently above the operator's head, making it difficult to monitor while simultaneously watching the road. Additionally, operators found it awkward to use the software button required to restart the shuttle movement after stops, due to its placement and their seating position.

\begin{figure}[htbp]
     \centering
     \begin{subfigure}[b]{0.42\textwidth}
         \centering
         \includegraphics[width=\textwidth, height=8cm]{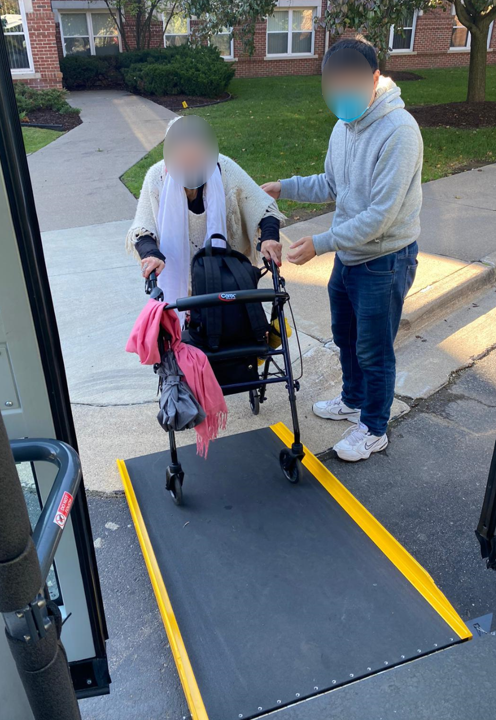}
         \caption{Operator helps a senior entering shuttle}
         \label{fig:seniors}
     \end{subfigure}
     \hfill
     \begin{subfigure}[b]{0.42\textwidth}
         \centering
         \includegraphics[width=\textwidth, height=8cm]{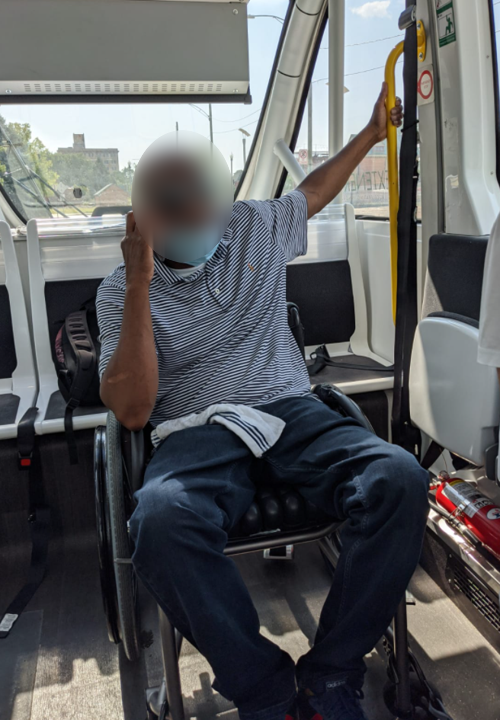}
         \caption{Hooks ensure the safety of disabled people}
         \label{fig:disabled}
     \end{subfigure}
        \caption{Seniors and Disabled Services}
        \label{fig:servingcustomers}
\end{figure}

Passenger feedback was mixed; while half were intrigued by the autonomous technology, others simply utilized the service for practical travel needs. The shuttle's autonomous capabilities were well-received when functioning correctly. However, the unexpected presence of an operator was a surprise to some, challenging their expectations of autonomous vehicle operations. Yet, for seniors and individuals with disabilities, the presence of an operator was deemed essential. To enhance accessibility, the manufacturer provided equipment such as ramps and wheelchair hooks as shown in Fig \ref{fig:servingcustomers}.  The ramp facilitated entry into the shuttle but its length, necessary for a gentle incline, limited its deployment at certain locations. While the hooks secured wheelchairs effectively during transit, passengers found them challenging to use independently, especially the hooks located behind them.

%% file: 4_planningRequirements.tex
\begin{figure}[htbp] 
\label{fig:bikes}
\centering
\includegraphics[width=1\linewidth]{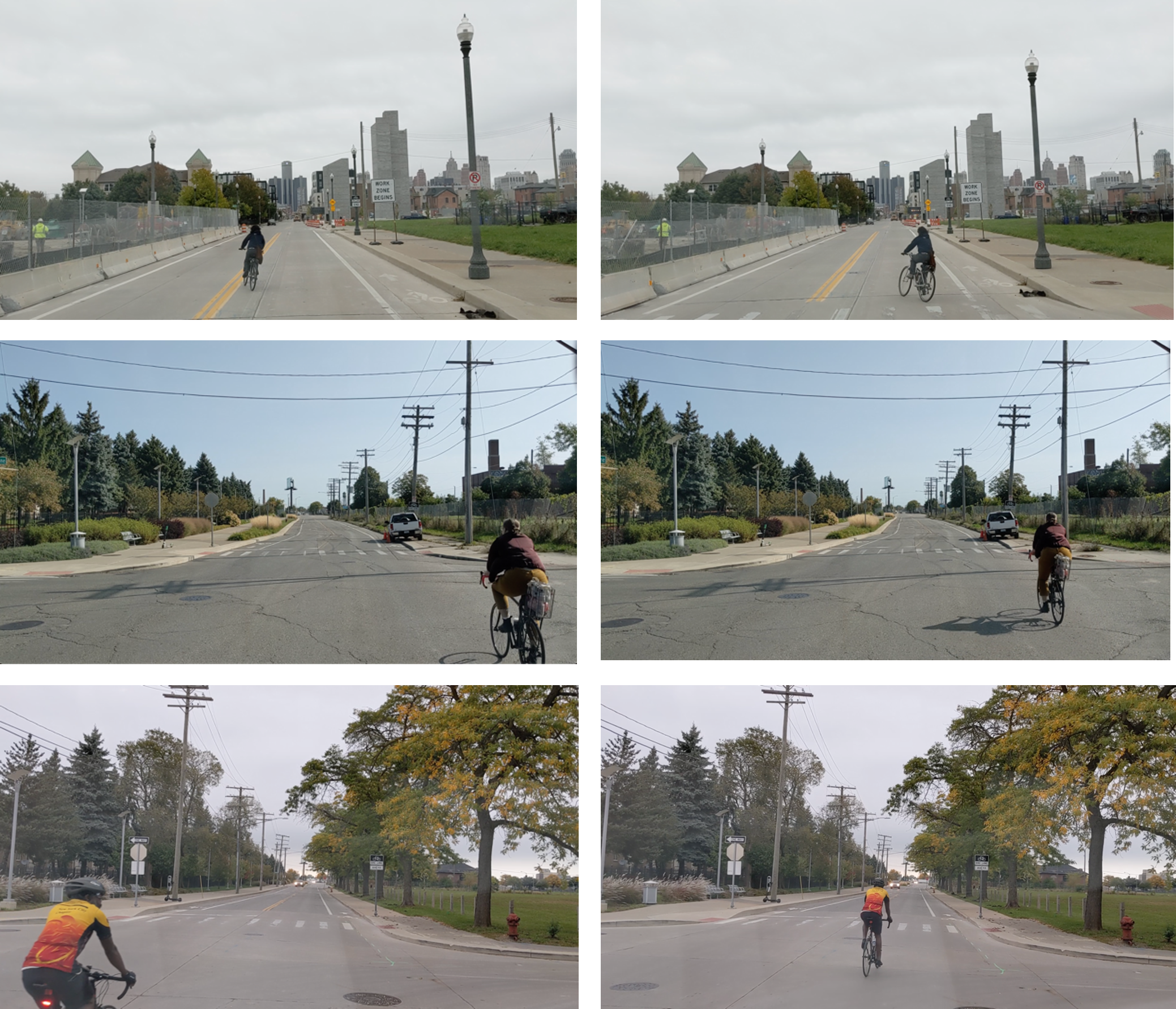} 
\caption{Bikes crossing the road, could cause unpredictable emergency brakes.} 
\label{collision} 
\end{figure}

\begin{figure}[htbp] 
\label{fig:steam}
\centering
\includegraphics[width=1\linewidth]{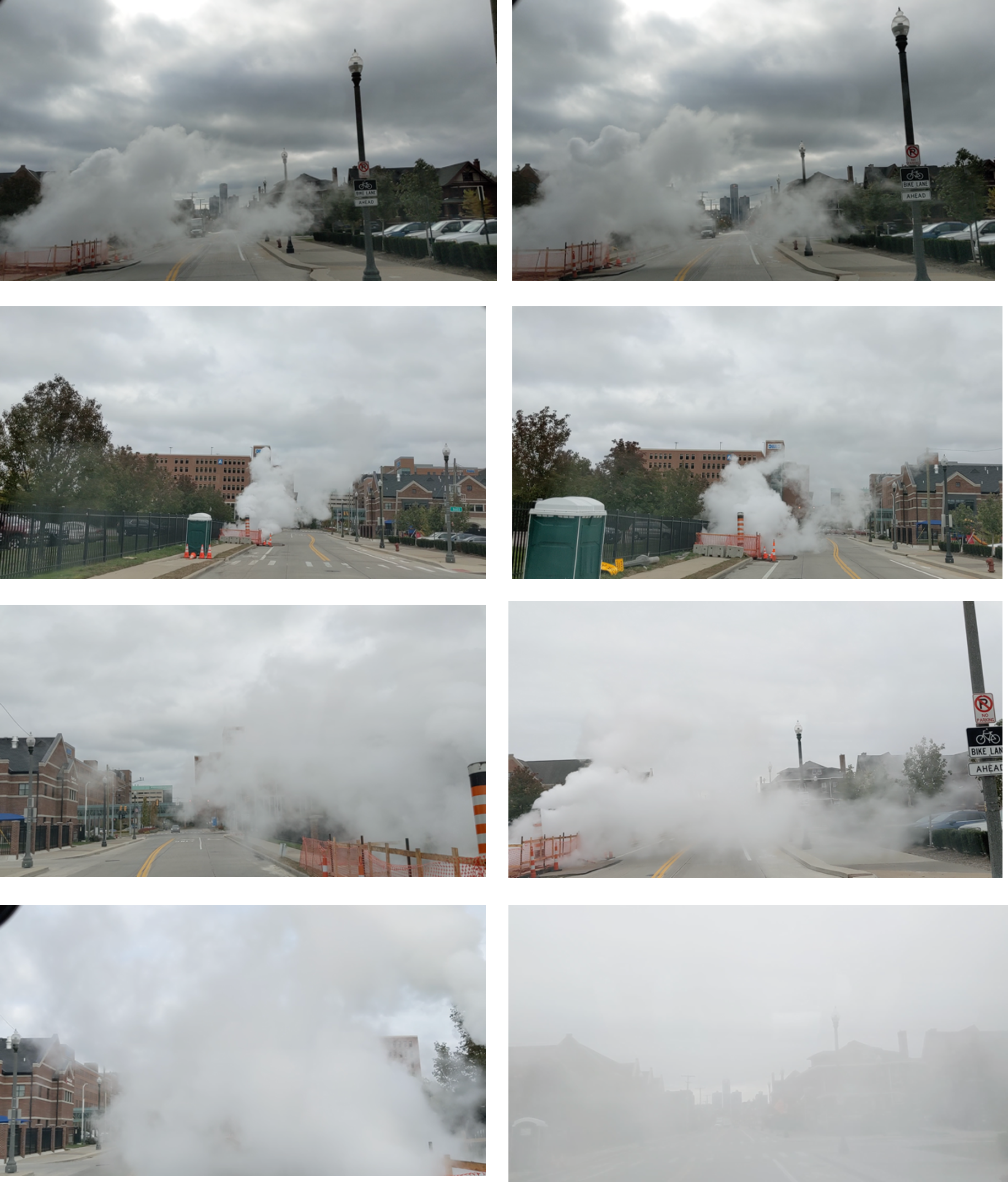} 
\caption{Steam on the street can block the Lidar reflection such that the car will recognize it as an obstacle, and conduct emergency brake.} 
\label{collision} 
\end{figure}

\begin{figure}[htbp] 
\label{fig:parking}
\centering
\includegraphics[width=1\linewidth]{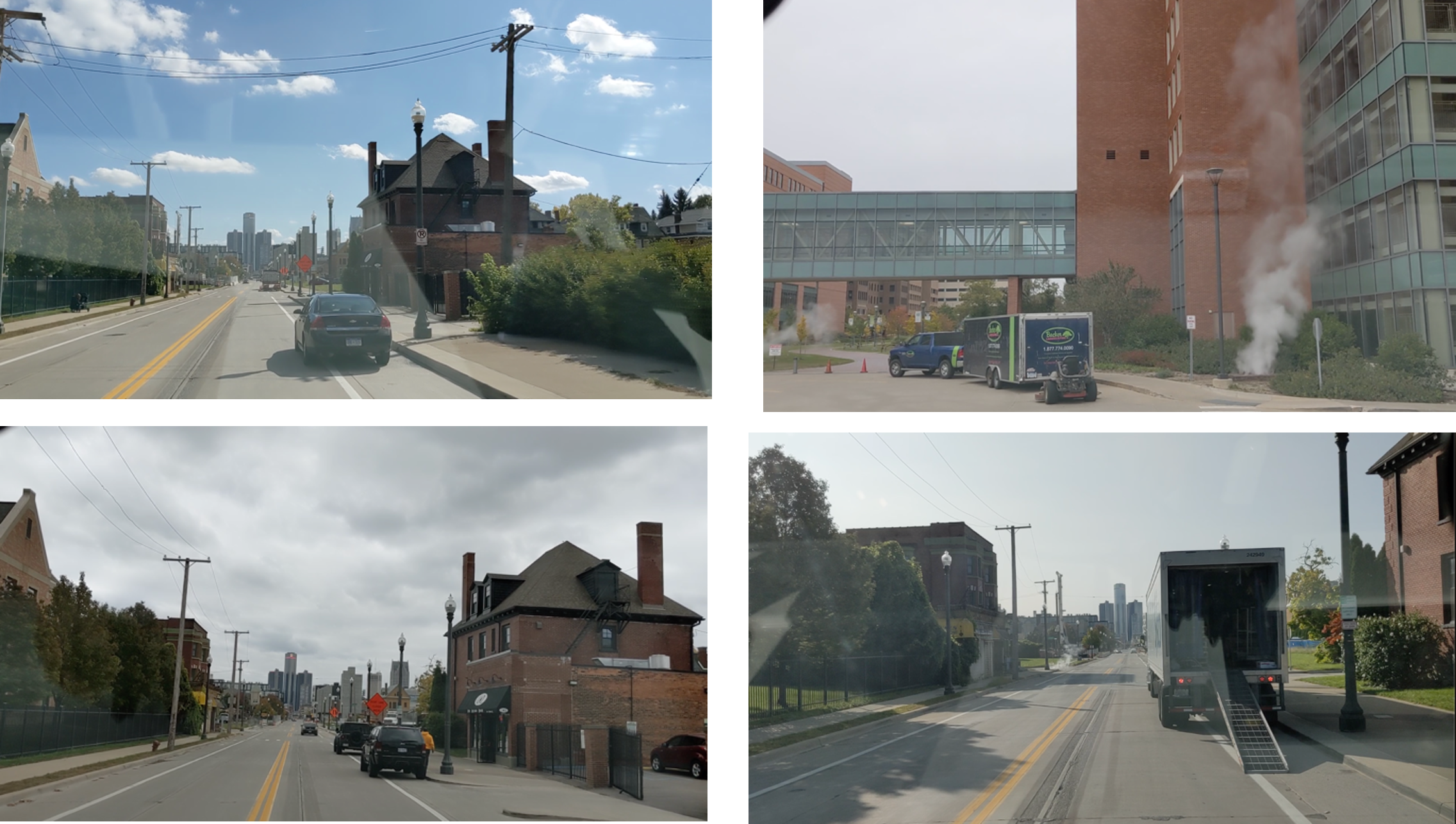} 
\caption{Parking cars occupied the lane.} 
\label{collision} 
\end{figure}

Based on the challenges identified during the operation of an autonomous shuttle in the Detroit area, we propose the following design and planning requirements to enhance the system's autonomy, safety, and passenger experience.

\subsection{Enhanced Localization and Decision-Making at Intersections}
\begin{itemize}
    \item \textbf{Requirement 1:} Implement advanced localization systems combining LiDAR, GPS, IMU, and V2X communications to improve accuracy at intersections, enabling the shuttle to make real-time decisions in complex scenarios.
    \item \textbf{Requirement 2:} Develop dynamic decision-making algorithms to predict the behavior of other vehicles and pedestrians accurately, allowing autonomous navigation through intersections without operator intervention.
\end{itemize}

\subsection{Dynamic Route Planning and Obstacle Avoidance}
\begin{itemize}
    \item \textbf{Requirement 3:} Introduce real-time dynamic route planning capabilities, enabling the shuttle to detect obstacles and reroute or navigate around them safely.
    \item \textbf{Requirement 4:} Upgrade sensor range and accuracy for enhanced obstacle detection and classification, ensuring safe navigation around unexpected obstacles and in narrow urban environments.
\end{itemize}

\subsection{Seamless Transition Between Manual and Autonomous Modes}
\begin{itemize}
    \item \textbf{Requirement 5:} Simplify the transition process for operators to resume autonomous mode, incorporating automated systems to assist in aligning the shuttle back on its route.
    \item \textbf{Requirement 6:} Implement fail-safe mechanisms for the shuttle to safely stop or navigate to a safe location if transitioning back to autonomous mode is not possible due to system failures.
\end{itemize}

\subsection{Operator Comfort and Efficiency}
\begin{itemize}
    \item \textbf{Requirement 7:} Redesign the operator’s workstation for improved ergonomics, including comfortable seating, accessible controls, and clear visibility of the operational dashboard and road.
    \item \textbf{Requirement 8:} Enhance the control terminal's user interface for intuitive display of critical information and ergonomic controls for frequently used functions.
\end{itemize}

\subsection{Passenger Experience and Accessibility}
\begin{itemize}
    \item \textbf{Requirement 9:} Refine and expand accessibility features to ensure easy access for all passengers, including those with disabilities, through easy-to-use ramps, adequate wheelchair space, and clear signage.
    \item \textbf{Requirement 10:} Implement an informative system inside the shuttle to educate passengers about autonomous technology and safety features, enhancing their comfort level and trust in the system.
\end{itemize}

Addressing these requirements will significantly improve the operational efficiency, safety, and passenger experience of autonomous shuttle services. By focusing on advanced localization, dynamic route planning, seamless transitions between driving modes, operator ergonomics, and passenger accessibility, the shuttle service can overcome current challenges and set a benchmark for future autonomous public transportation systems.

%% file: 5_Discussion.tex
In the previous section, we presented the safety issues observed during testing and the operators' and passengers' experiences of riding the autonomous shuttle. These safety issues caused by shortages of autonomous driving systems and unexpected road events demonstrate the challenge of designing the autonomous driving system. In this section, we will propose two initiatives for enhancing autonomous vehicles which are concluded from our observations.

\subsection{Requirement of Teleoperation system}
Tele-operation, or remote operation, has been implemented to work with many robotics systems and mobile robots. A disaster rescuer can control a vehicle by diving into the disaster zone to search for survivors. Doctors can conduct remote surgery over long distances. Drones enable the operator to experience an immersive, high-speed flying experience with a virtual reality helmet. By analyzing the existing teleoperation system, we notice that all systems can be described with a diagram. The diagram includes a robot that can be controlled with an electronic signal, uplink, and downlink communication channels that are responsible for uploading feedback to the operator and sending a command to the robot, respectively, with low and stable latency, and a human-machine interface for the operator to understand the remote conditions and effectively generate an operation command that meets his intention. At this point, vehicle teleoperation is still in its initial stages. Compared to any other teleoperation application, vehicle teleoperations are hard to implement for three challenges: 1) The conditions of the road are commonly complex for an operator to make a decision; 2) vehicles are more latency-sensitive than control, which is delayed by half a second and may cause traffic incidents; and 3) the effective range of vehicle teleoperation must be large enough to be useful. On the other hand, autonomous vehicles have three advantages for implementing teleoperation: 1) Various high-definition sensors are equipped on vehicles; 2) powerful computing platforms enable intelligent driving assistance applications running on vehicles; and 3) wireless cellular networks have been significantly improved. 

\subsection{Initiative for improving Accessibility for Visual disabled people }\label{AA}
In recent years, various transportation and mobility options have tightly involved people's lives and benefited most people in their day-to-day activities. However, many mobility options are still inconvenient and physically inaccessible for people with disabilities. In the United States, approximately one in every five people, or more than 57 million people, has a disability \cite{disabilityRights}, including more than 3.8 million veterans with a service-connected disability. Besides, approximately 12 million people aged 40 and over in the United States have a vision impairment, including one million who are blind. Among these people with disabilities, six million individuals have difficulty getting the transportation they need. Transportation is key community-based support that allows individuals to fulfill their civic responsibility and makes it possible to enjoy a quality of life. When a disability limits transportation options, this can result in reduced economic opportunities, isolation that exacerbates medical conditions or leads to depression, and a diminished quality of life. 

Generally speaking, mobility can be defined as traveling from point A to point B. But how does an individual with a disability navigate to or identify Point A What happens after arriving at Point B? One of mobility’s toughest challenges for an individual with a disability is providing a safe and simple solution to the First Mile/Last Mile problem. People who have visual disabilities struggle to travel that short distance to or from a bus stop and other points of interest due to their physical limitations. In this Mobility-for-All challenge, our group is focusing on dealing with the problem of how to provide blind people with safe and stable navigation in the first mile or last mile. In this case, we propose a product ''Guardian Angel'', which provides blind people navigation to help locate and identify pick-up and drop-off locations and building entrances/exits. This user-friendly product could eliminate the First Mile/Last Mile problem and offer true point-to-point, demand-response service.

\subsection{Initiative for Autonomous Planning and Online Decision Making}

This initiative aims to advance the shuttle's capability in real-time planning, decision-making, and obstacle avoidance. It focuses on developing sophisticated algorithms, sensor fusion, and machine-learning techniques for enhanced autonomy and safety. The objective is to create a robust sensor fusion system for precise localization and to implement machine learning algorithms for dynamic decision-making. V2X communication will increase situational awareness, and advanced obstacle detection algorithms will enable real-time navigation around obstacles. Online learning mechanisms will allow the shuttle to adapt its decision-making processes over time, improving performance in navigating urban environments. An operator-assisted control system will provide decision support, facilitating seamless transitions between autonomous and manual control.

%% file: 6_Summary.tex
Due to the driver shortage problem, public transportation is facing horrible job vacancies in the near future, leading to challenges in accessing health services and social events. Seniors and disabled people are actually becoming the group that requires autonomous vehicles the most. Through the deployment of autonomous shuttles for seniors and the disabled, we have observed quite a lot of problems with the current autonomous driving system that are extremely unfriendly to its target passengers. Starting from the more similar small-scale deployment can facilitate the development of autonomous vehicles by solving the edge cases in the real world and improving social acceptance with real benefits supplied by the emerging technology.

%% file: 7_Acknowledgement.tex
The authors wish to extend their sincere gratitude to Ray Smith from IXR Mobility for the opportunity to engage in the shuttle operation. This experience has been invaluable to our research and practical understanding in the autonomous driving field. We also express our appreciation to Navya Tech for providing essential training to our operators, which has been crucial for ensuring the safe implementation of our project activities.